\documentclass[sigconf]{acmart}

\usepackage{balance}
\usepackage{hyperref}
\usepackage{latexsym}
\usepackage{graphicx}
\graphicspath{{./images/}}
\usepackage{booktabs} 
\usepackage{color}  
\usepackage{amsmath}  
\usepackage{subcaption}
\usepackage{caption}
\usepackage{tikz}
\usepackage{colortbl} 
\usepackage{framed}
\usepackage{multirow}
\usepackage{multicol}
\usepackage{url}
\usepackage{verbatim}
\usepackage{cancel}
\usepackage{xspace} 
\usepackage[ruled,linesnumbered,vlined]{algorithm2e}
\usepackage{bbold} 
\usepackage{arydshln}
\usepackage{float}

\newcommand{\utterance}[1]{\textit{#1}}
\newcommand{\phrase}[1]{\textit{``#1''}}

\newcommand{\myparagraph}[1]{\noindent \textbf{#1}.}

\newcommand{\squishlist}{
	\begin{list}{$\bullet$}
		{ \setlength{\itemsep}{0pt}
			\setlength{\parsep}{3pt}
			\setlength{\topsep}{3pt}
			\setlength{\partopsep}{0pt}
			\setlength{\leftmargin}{1.5em}
			\setlength{\labelwidth}{1em}
			\setlength{\labelsep}{0.5em} } }
	\newcommand{\squishend}{
\end{list}  }

\newcommand{\ragonite}{\textsc{RAGonite}\xspace}
\newcommand{\confquestions}{\textsc{ConfQuestions}\xspace}

\AtBeginDocument{%
  \providecommand\BibTeX{{%
    \normalfont B\kern-0.5em{\scshape i\kern-0.25em b}\kern-0.8em\TeX}}}

\providecommand\BibTeX{{
 Bib\TeX}}

\copyrightyear{2025}
\acmYear{2025}
\setcopyright{rightsretained}
\acmConference[WSDM '25]{Proceedings of the Eighteenth ACM International Conference on Web Search and Data Mining}{March 10--14, 2025}{Hannover, Germany}
\acmBooktitle{Proceedings of the Eighteenth ACM International Conference on Web Search and Data Mining (WSDM '25), March 10--14, 2025, Hannover, Germany}\acmDOI{10.1145/3701551.3704126}
\acmISBN{979-8-4007-1329-3/25/03}

\makeatletter
\gdef\@copyrightpermission{
 \begin{minipage}{0.3\columnwidth}
  \href{https://creativecommons.org/licenses/by/4.0/}{\includegraphics[width=0.90\textwidth]{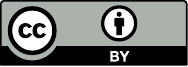}}
 \end{minipage}\hfill
 \begin{minipage}{0.7\columnwidth}
  \href{https://creativecommons.org/licenses/by/4.0/}{This work is licensed under a Creative Commons Attribution International 4.0 License.}
 \end{minipage}
 \vspace{5pt}
}
\makeatother

\begin{document}


\title[Evidence Contextualization and Counterfactual Attribution for ConvQA over Heterogeneous  Data with RAG Systems]{Evidence Contextualization and Counterfactual Attribution for Conversational QA over Heterogeneous Data with RAG Systems}

\author{Rishiraj Saha Roy}
\orcid{0000-0002-5774-5658}
\affiliation{
    \department{Audio and Media Technologies}
    \institution{Fraunhofer IIS}
    \streetaddress{Am Wolfsmantel 33}
    \city{Erlangen}
    \state{Bavaria}
    \postcode{91058}
    \country{Germany}}
\email{rishiraj.saha.roy@iis.fraunhofer.de}

\author{Joel Schlotthauer}
\orcid{0009-0009-6525-6094}
\affiliation{
    \department{Audio and Media Technologies}
    \institution{Fraunhofer IIS}
    \streetaddress{Am Wolfsmantel 33}
    \city{Erlangen}
    \state{Bavaria}
    \postcode{91058}
    \country{Germany}}
\email{joel.schlotthauer@iis.fraunhofer.de}

\author{Chris Hinze}
\orcid{0009-0006-3196-6280}
\affiliation{
    \department{Audio and Media Technologies}
    \institution{Fraunhofer IIS}
    \streetaddress{Am Wolfsmantel 33}
    \city{Erlangen}
    \state{Bavaria}
    \postcode{91058}
    \country{Germany}}
\email{chris.hinze@iis.fraunhofer.de}

\author{Andreas Foltyn}
\orcid{0009-0001-4678-9786}
\affiliation{
    \department{Audio and Media Technologies}
    \institution{Fraunhofer IIS}
    \streetaddress{Am Wolfsmantel 33}
    \city{Erlangen}
    \state{Bavaria}
    \postcode{91058}
    \country{Germany}}
\email{andreas.foltyn@iis.fraunhofer.de}

\author{Luzian Hahn}
\orcid{0009-0005-8450-2714}
\affiliation{
    \department{Audio and Media Technologies}
    \institution{Fraunhofer IIS}
    \streetaddress{Am Wolfsmantel 33}
    \city{Erlangen}
    \state{Bavaria}
    \postcode{91058}
    \country{Germany}}
\email{luzian.hahn@iis.fraunhofer.de}

\author{Fabian Kuech}
\orcid{0009-0005-4665-387X}
\affiliation{
    \department{Audio and Media Technologies}
    \institution{Fraunhofer IIS}
    \streetaddress{Am Wolfsmantel 33}
    \city{Erlangen}
    \state{Bavaria}
    \postcode{91058}
    \country{Germany}}
\email{fabian.kuech@iis.fraunhofer.de}

\renewcommand{\shortauthors}{Roy et al.}


\begin{abstract}
Retrieval Augmented Generation (RAG) works as a backbone for interacting with an enterprise's own data via Conversational Question Answering (ConvQA). In a RAG system, a retriever fetches passages from a collection in response to a question, which are then included in the prompt of a large language model (LLM) for generating a natural language (NL) answer. However, several RAG systems today suffer from two shortcomings: (i) retrieved passages usually contain their raw text and lack appropriate document context, negatively impacting both retrieval and answering quality; and (ii) attribution strategies that explain answer generation typically rely only on similarity between the answer and the retrieved passages, thereby only generating plausible but not causal explanations. In this work, we demonstrate \ragonite, a RAG system that remedies the above concerns by: (i) contextualizing evidence with source metadata and surrounding text; and (ii) computing counterfactual attribution, a causal explanation approach where the contribution of an evidence to an answer is determined by the similarity of the original response to the answer obtained by removing that evidence. To evaluate our proposals, we release a new benchmark \confquestions: it has $300$ hand-created conversational questions, each in English and German, coupled with ground truth URLs, completed questions, and answers from $215$ public Confluence pages. These documents are typical of enterprise wiki spaces with heterogeneous elements. Experiments with \ragonite on \confquestions show the viability of our ideas: contextualization improves RAG performance, and counterfactual explanations outperform standard attribution.
\end{abstract}

\settopmatter{printacmref=true, printccs=true, printfolios=true}



\maketitle

\section{Introduction}
\label{sec:intro}

\myparagraph{Motivation} ``Talk to your data'' is a major research theme today, where users interact with local knowledge repositories to satisfy their information needs. \textit{Conversational question answering (ConvQA)}~\cite{reddy2019coqa,christmann2019look} is a natural choice for such interactions, where a user starts with a self-sufficient (\textit{intent-explicit}) question, and follows that up with more ad hoc, conversational questions that leave parts of the context unspecified (\textit{intent-implicit}). Hand in hand, the emergence of powerful large language models (LLMs) has led to \textit{retrieval augmented generation (RAG)}~\cite{lewis2020retrieval} as the backbone for designing QA systems over one's own data. In RAG pipelines, given a \textit{question}, a retriever fetches relevant evidence from local data, and this is passed on to an LLM for a concise and fluent \textit{answer} to the user's question. Enterprise knowledge repositories usually consist of documents with \textit{heterogeneous} elements, i.e. they often contain interleaved tables, lists and passages (henceforth, we use the unifying term \textit{evidence}). Typical examples of such documents with mixed structured and unstructured elements are meeting notes, test reports, or product descriptions. The setting for this demo is \textit{ConvQA} with \textit{RAG} over such \textit{heterogeneous} document collections~\cite{pan2022trag,yu2022retrieval}.

\myparagraph{Limitations of state-of-the-art} Over the last four years, RAG has been a topic of intense investigation~\cite{gao2023retrieval}. Beyond literature, organizations like LangChain, LlamaIndex, or Cohere offer frameworks to build RAG systems. We posit that despite several advanced features, these suffer from two basic concerns, one at each end of the pipeline: (i) at the beginning, documents are typically split into chunks (usually one or more passages) that are indexed on their content~\cite{gao2023retrieval}. When these chunks are retrieved and fed to an LLM, they often lack supporting \textit{context} from the document, which adversely affects both retrieval, and subsequent answering; and (ii) at the end, \textit{attribution} mechanisms that attach provenance likelihoods of the answer to the retrieved units of evidence, are solely based on statistical similarity between the answer and the evidence~\cite{li2023survey}: these are not causal~\cite{pearl1997causation}, but rather only \textit{plausible} explanations.
Moreover, current pipelines support raw text and rarely mention how tabular elements could be handled.

\begin{figure} [t]
	\centering
	\includegraphics[width=\columnwidth]{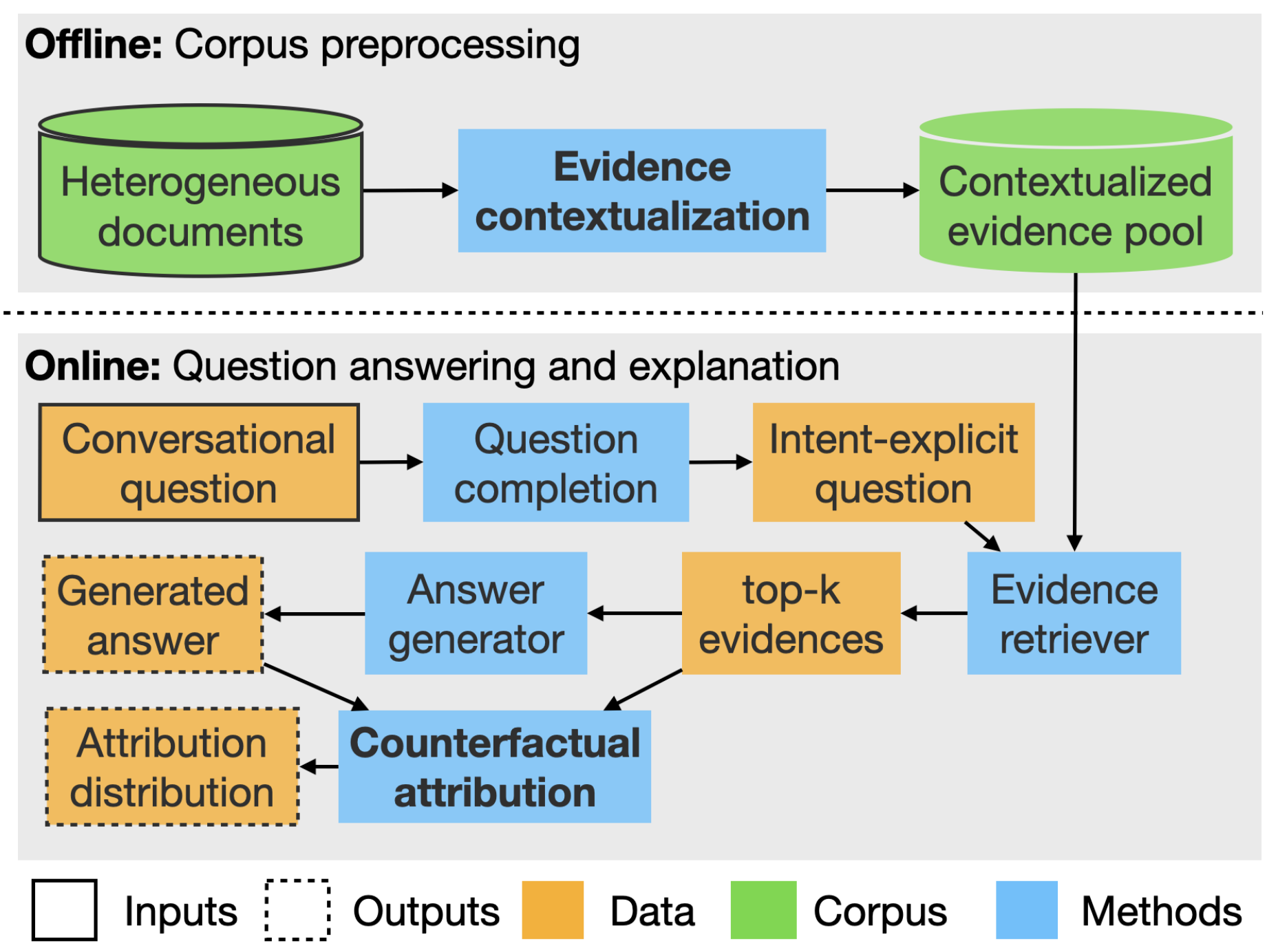}
    \vspace*{-0.7cm}
	\caption{The \ragonite workflow enhances RAG pipelines at both ends, preprocessing evidence and explaining answers.} 
	\label{fig:overview}
    \vspace*{-0.5cm}
\end{figure}

\myparagraph{Contributions} We make the following salient contributions:
\squishlist
    \item We demonstrate \ragonite,
    a new RAG system that concatenates page titles, headings, and surrounding text to raw contents of evidences for better retrieval and answering;
    \item We compute counterfactual attribution distributions over retrieved evidences as causal explanations for answers;
    \item We bring tables under \ragonite's scope by linearizing each record (row) via verbalization~\cite{oguz2022unik} and similar techniques~\cite{li2023table};
    \item We create \confquestions, a benchmark with 
    $300$ conversational questions for evaluating RAG-based heterogeneous QA. 
\squishend

The benchmark and all other artifacts for this work are public at
\url{https://github.com/Fraunhofer-IIS/RAGonite}.

\begin{figure*} [t]
	\centering
    \includegraphics[width=\textwidth]{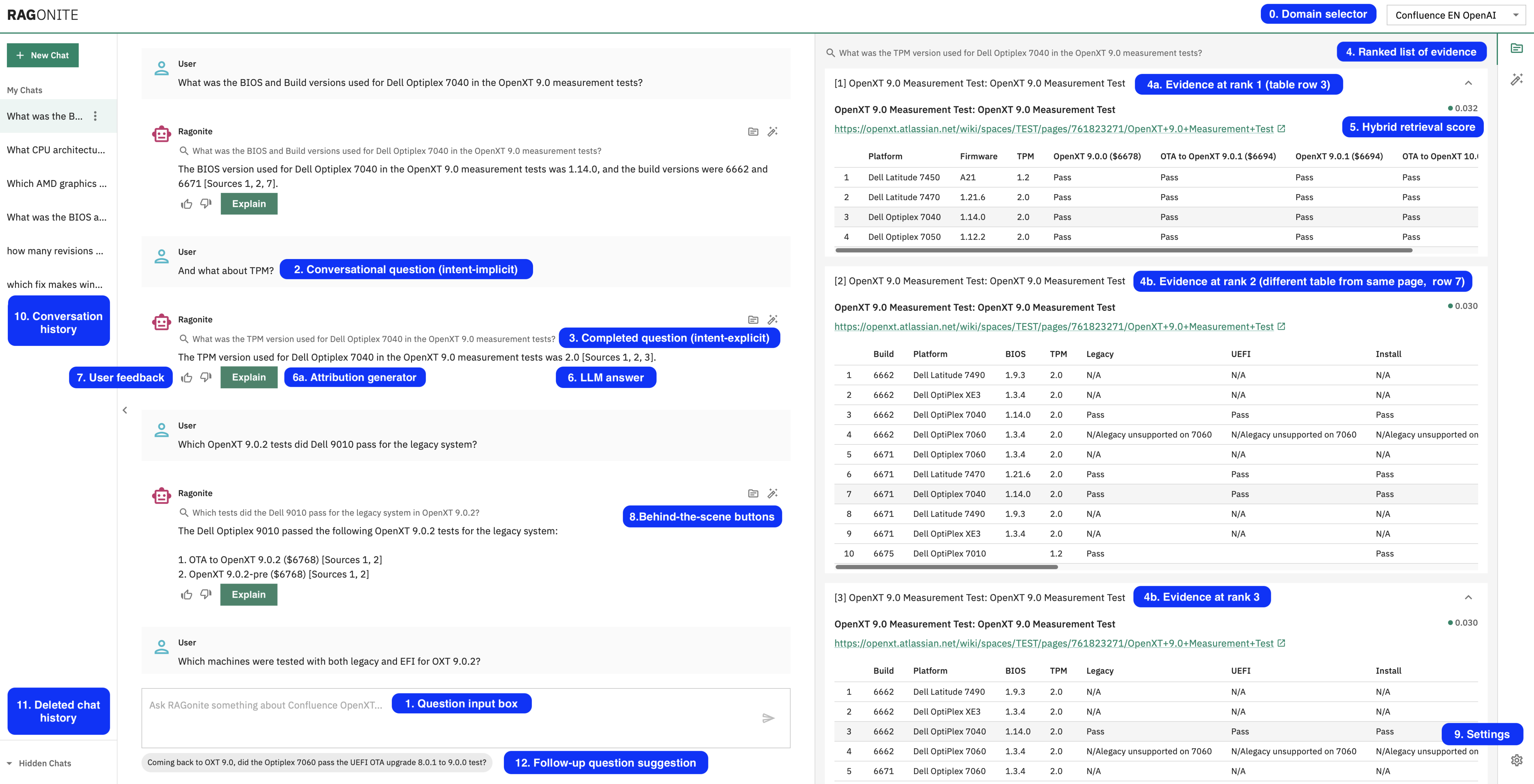}
	\vspace*{-0.7cm}
	\caption{An annotated walkthrough the of \ragonite demo. Blue boxes guide the reader and are not part of the UI (Sec.~\ref{sec:walkthrough}).}
	\label{fig:screenshot}
    \vspace*{-0.3cm}
\end{figure*}

\section{System overview} 
\label{sec:overview}

\myparagraph{Backend}
An overview of the \ragonite pipeline is in Fig.~\ref{fig:overview}.
RAGonite's backend is split into a functional core, which handles retrieval and answer generation, and a stateful layer that persists chats into a SQLite database and provides a REST API to the frontend using FastAPI.
The dependencies in the functional core include the vector database (ChromaDB), and prompt template (Jinja) and LLM libraries (\texttt{gpt-4o} and \texttt{Llama-3.1-8B}).
We use ChromaDB as our vector database for storing the contextualized evidences, and also use ChromaDB's in-built retrieval functions.
The multilingual BGE embedding model\footnote{\url{https://huggingface.co/BAAI/bge-m3}} was used to embed evidences, that was found to work slightly better than \texttt{text-embedding-3-small}\footnote{\url{https://openai.com/index/new-embedding-models-and-api-updates/}} from OpenAI.
While ChromaDB was used for its extensibility, we also explored variants like Weaviate and Milvus.
We used top-k hybrid search with 
the BGE reranker\footnote{\url{https://huggingface.co/BAAI/bge-reranker-v2-m3}} and reciprocal rank fusion (RRF)~\cite{cormack2009reciprocal} being used to merge $k$ dense retrieval and lexical retrieval results.
We use GPT-4o as our LLM of choice inside the question completion and answer generation modules for efficiency and quality, but we also support a Llama model in our demo.
All prompts can be seen inside the demo interface for transparency.
A single GPU server (4x48GB NVIDIA Ada 6000 RTX, 512 GB RAM, 64 virtual cores) was used for all our experiments. All code is in Python. 

\myparagraph{Frontend}
For the frontend we developed a single-page React application (actually Preact, as a more lightweight React-compatible alternative), intentionally avoiding additional dependencies to eliminate the need for a build process. All API calls in the \ragonite demonstration are handled by the frontend.

\section{Demo walkthrough}
\label{sec:walkthrough}

\begin{figure} [t]
	\centering
    \includegraphics[width=\columnwidth]{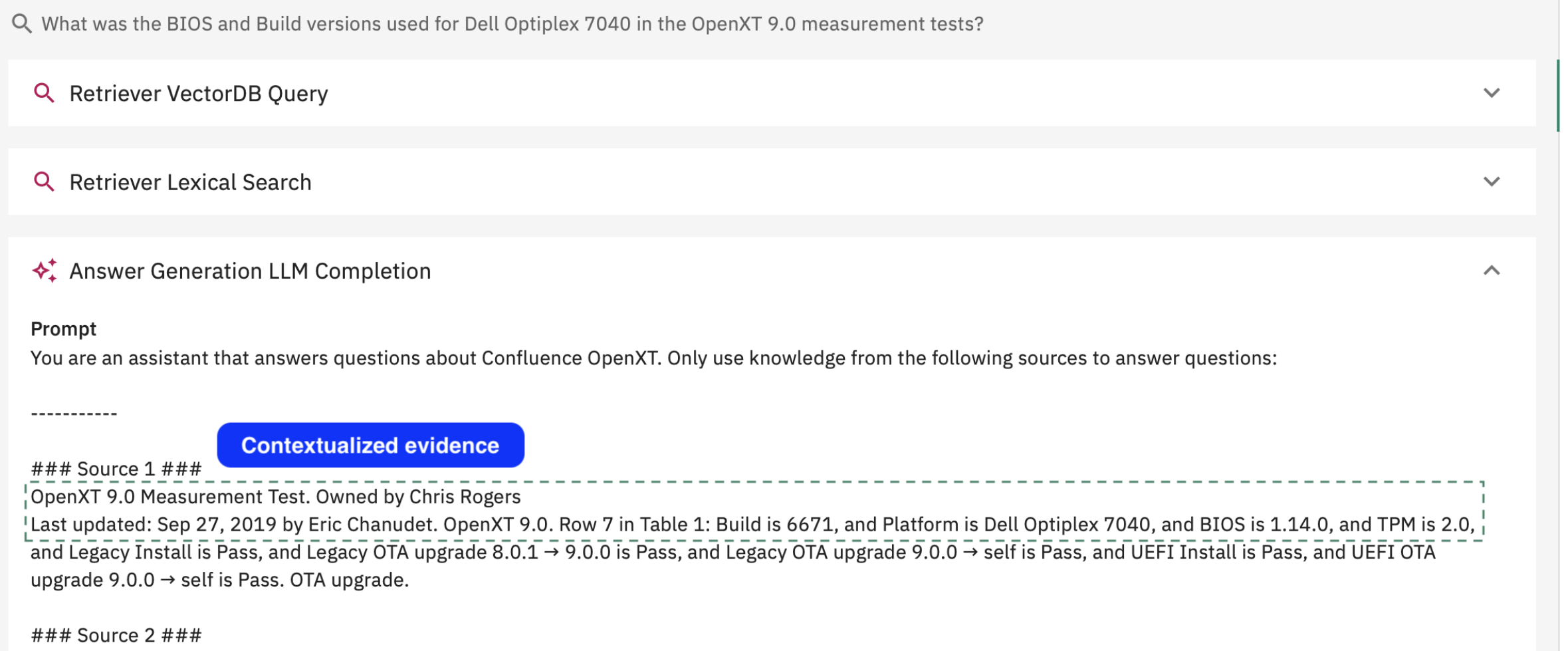}
	\vspace*{-0.7cm}
	\caption{Contextualized evidence as in answer prompt.}
	\label{fig:transparency}
    \vspace*{-0.3cm}
\end{figure}

\begin{figure} [t]
	\centering
    \includegraphics[width=\columnwidth]{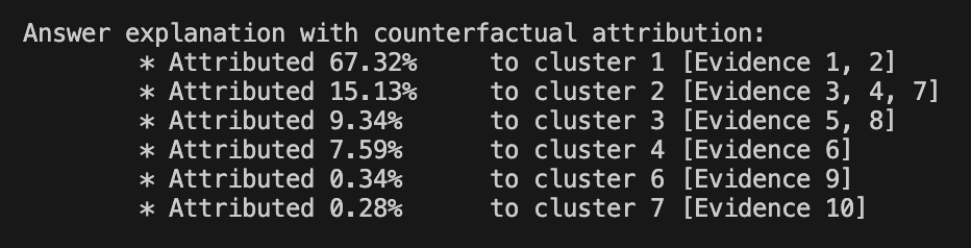}
	\vspace*{-0.7cm}
	\caption{Answer explanation by counterfactual attribution.}
	\label{fig:attrib}
    \vspace*{-0.5cm}
\end{figure}

We use a screenshot of the \ragonite main page in Fig.~\ref{fig:screenshot}, on which we overlay numbered blue boxes, for a demo walkthrough. First, a user must select a domain (0) on which to use \ragonite (our focus here is on enterprise wikis exemplified by Confluence, but \ragonite also runs on other domains like soccer, automobiles, and fictional universes). Then the user uses the question input box (1) to begin a conversation. Suppose we are at turn two (2):
when \ragonite receives a conversational question as input, it uses its \textit{question completion} module to rephrase the question into an intent-explicit form using an LLM (3). While the question in the first conversation turn is usually self-contained, follow-up questions are completed using relevant information from \textit{previous questions and generated answers}. Offline, the heterogeneous document collection is preprocessed via our evidence contextualization module (Sec.~\ref{sec:context}) into a pool of evidence where necessary document context is concatenated to the raw contents of the evidence
(example in Fig.~\ref{fig:transparency}, where the page title and the preceding heading is prepended to a verbalized table record).
A \textit{retriever} then takes the intent-explicit question and searches over the evidence pool, to return top-k question-relevant evidences (4, 4a-c) using the hybrid retrieval score (5). These top-k evidences are inserted into the prompt of another LLM instance to generate the answer (6). Finally, the generated answer and the top-k evidences are fed into our counterfactual attribution module (Sec.~\ref{sec:attribute}). When a user clicks on ``Explain'' (6a), a command line window pops up to output the attribution distribution as an explanation of how the answer was potentially constructed (Fig.~\ref{fig:attrib}). We log user feedback on the answer for future use (7). A user can see a trace through the pipeline using behind-the-scenes buttons (8): for example, retrieval results for lexical and dense search, LLM prompts, and more. Users can adjust retriever and generator configurations (9), see their past conversations (10), including deleted ones (11).
Follow-up question suggestions are also provided (12) to help domain exploration.
Notably, most open-source RAG demos only consist of the evidence retriever and answer generator modules.
On average, \ragonite requires about one second to answer a question, and approximately two seconds for explaining the answer.

\section{Evidence contextualization}
\label{sec:context}

\begin{figure} [t]
	\centering
    \includegraphics[width=\columnwidth]{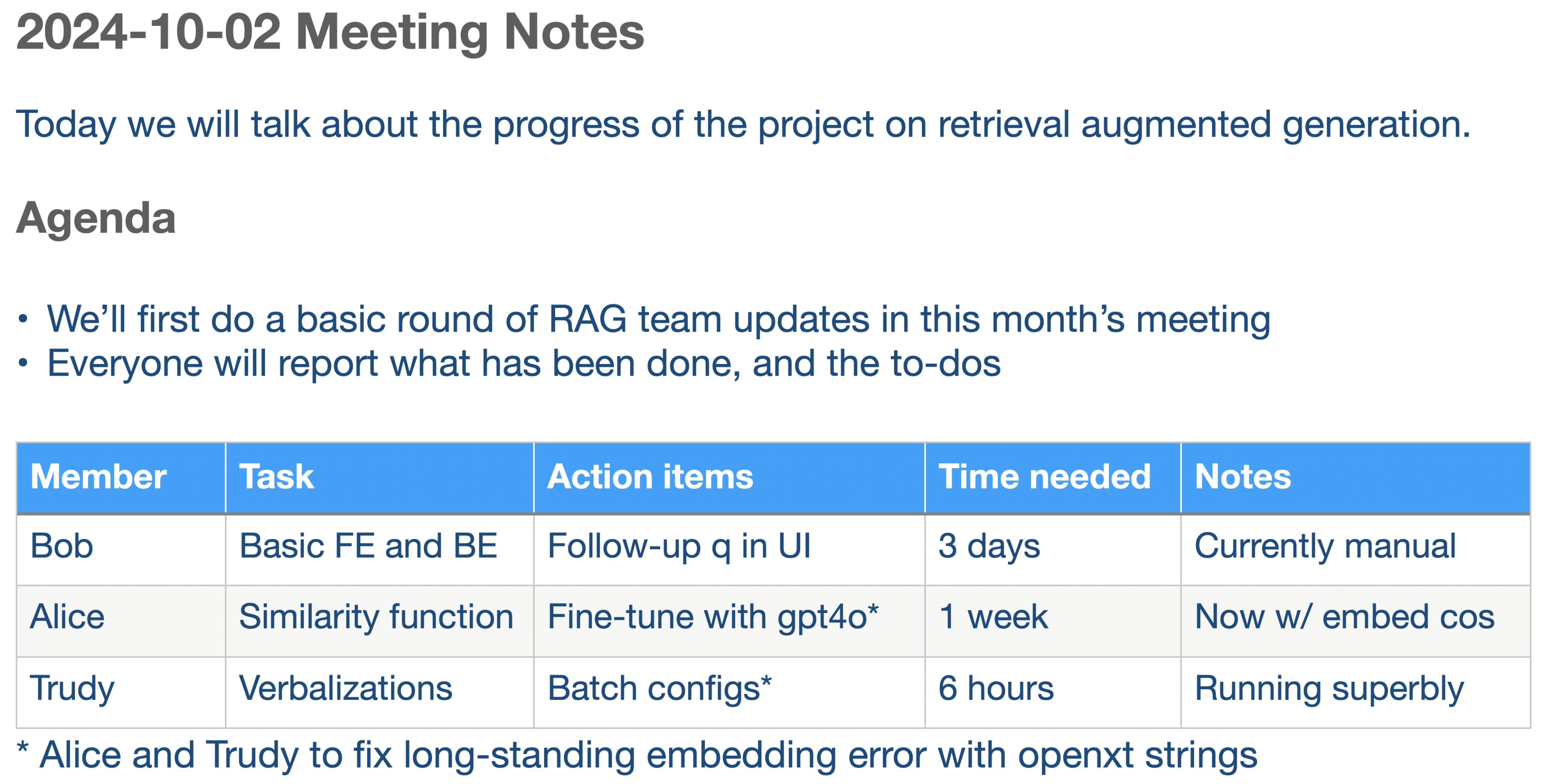}
	\vspace*{-0.7cm}
	\caption{Toy wiki page with heterogeneous elements to motivate evidence contextualization. A question like \utterance{todo for alice in oct rag meeting?} can only be faithfully answered by joining information in the relevant table row, table footer, page title, and preceding heading and text. This implies that unless the evidence as stored in the DB contains the supporting context, there is no chance that a retriever can fetch it from the corpus with the question as a search query.}
	\label{fig:webpage}
    \vspace*{-0.3cm}
\end{figure}

\myparagraph{Idea} Evidence contextualization in current RAG systems mainly involve steps like: (i) coreference resolution for resolving simple pronouns like (s)he, her, his, etc., and (ii) overlapping text chunking involving sliding windows over passages~\cite{gao2023retrieval}. These have some basic limitations like: (i) resolving pronouns is mainly limited to previous entity references, and is not enough for more subtle coreferences like \phrase{these configurations} or \phrase{the previous model}; (ii) sliding word/token windows do not generalize well to structured evidence like tables. We, however, adopt a simple but effective alternative 
of concatenating document context to each evidence
at indexing time. 

A toy example of an enterprise wiki page is in Fig.~\ref{fig:webpage}.
We begin by capturing text inside \texttt{<table>...</table>} tags as tables. Text inside \texttt{<ol>...</ol>} (ordered list) and \texttt{<ul>...</ul>} (unordered list) tags are stored as lists. Each span of the remaining text between any \texttt{<heading>}-s, or between a \texttt{<heading>} and the beginning or end of the document, is assigned to one passage.
Each list and each paragraph become individual pieces of evidence.

\myparagraph{Preprocessing} We store each table in a verbalized form~\cite{oguz2022unik}, which converts structured evidence to a form more amenable to an LLM prompt while retaining scrutability by a human. In this mode, we traverse left to right in a table row, and linearize the content as ``<Column header 1> is <value 1>, and <Column header 2> is <value 2>, ...''. We prepend this text with ``Row <id> in Table <id>'' to preserve original row ordering information as well as table ordering relative to the page where the content is located.
For instance, the second row in the table in Fig.~\ref{fig:webpage} would be verbalized as \phrase{Row 2 in Table 1: Member is Alice, and Task is Similarity function, and ...}.
Along with the complete table, each linearized row is also stored as an individual evidence. This helps pinpoint answer provenance to individual rows in tables when possible, as well as resolve comparative questions that are easier when the whole table is in one piece in the prompt.

\myparagraph{Contextualization} At this point, we have passages, lists, tables, and table records as individual evidences. We then concatenate the following items to each piece of evidence: page title, previous heading, the evidence before, and the evidence after. Both the raw textual forms and their embeddings are indexed in our database, enabling lexical and dense search, respectively.
Note that indexing whole pages and inserting full text of top-k documents into an LLM prompt cannot easily bypass evidence contextualization, due to arbitrary lengths of web documents: operating at evidence-level is therefore a practical choice.

An example of contextualized evidence is in Fig.~\ref{fig:transparency} (inside dashed box): the raw content of the evidence (the seventh row in the first table) starts with \phrase{Row 7 in Table 1...} But the previous content starting at \phrase{OpenXT 9.0 ...} consists of the page title, preceding heading, and preceding text. \phrase{OTA upgrade} at the end comes from adding succeeding text (the table footer here).

\vspace*{-0.2cm}
\section{Counterfactual attribution}
\label{sec:attribute}

\setlength{\textfloatsep}{0pt} 
\begin{algorithm}[t]
    \small
	\textbf{Input:} Question $q$, Evidences $E = \{e\}$, Answer $a$, MC iterations $m$\\
	\textbf{Output:} Distribution $\boldsymbol{\mathcal{A}}$ for attributing answer $a$ to each evidence $e$\\
	\SetAlgoLined
	\DontPrintSemicolon
	$E^{cl} = \{e^{cl}_i\} \leftarrow Cluster(E)$ \tcp*{Group redundant evidences}
	\For(\tcp*[f]{For each evidence cluster}){$e^{cl}_i \in E^{cl}$} {
		$E^{cl, cf}_i \leftarrow E^{cl} \setminus e^{cl}_i$ \tcp*{Create counterfactual evidence}
        \For(\tcp*[f]{Run Monte Carlo iterations}){$j \in 1\ldots m$} {
			$a^{cf}_{i,j} = LLM(q, E^{cl, cf}_i)$ \tcp*{Generate cf answer}
            \textbf{compute} $s_{i,j} \leftarrow sim(a, a^{cf}_{i,j})$ \tcp*{Compute similarity}
        }
        \textbf{compute} $c_i \leftarrow 1 - \sum_j s_{i,j} / m$ \tcp*{Contribution of $e^{cl}_i$ to a}
        \textbf{compute} $\boldsymbol{\mathcal{A}} \leftarrow softmax(c_i) \forall e^{cl}_i$ \tcp*[l]{Normalize to $[0,1]$}
    }
	\Return $\boldsymbol{\mathcal{A}}$
	\caption{Counterfactual attribution in \ragonite}
	\label{algo:cf-attrib}
\end{algorithm}

\myparagraph{Idea} Counterfactual explanations~\cite{tran2021counterfactual} through passage perturbation has been explored in the context of RAG~\cite{sudhi2024rag} and more generally for LLM understanding~\cite{xi2024eadaptive}, but not for answer attribution and not via evidence removal, as explore in this work.
The intuition here is that a definitive way of saying how much an evidence $e_i$ \textit{contributed} to an answer $a$ is to \textit{remove} the evidence from the prompt, and see how much the original answer $a$ \textit{changed}.
If $a$ did not change much, then $e_i$ is unlikely to have played any major role, and vice versa.

\myparagraph{Counterfactual Attribution (CFA)} This simple idea is the core of our proposed Algorithm~\ref{algo:cf-attrib}. The only other confounding factor is the presence of \textit{redundant evidences} in the prompt: the retriever may sometimes fetch two (or more) semantically equivalent evidences if they both satisfy the user's intent. In the context of counterfactuals, it may be that when one of these redundant evidences are removed, the answer hardly changes. This may lead to the wrong conclusion that the first evidence was unimportant: it just happens that the second evidence makes up for it content-wise. Hence, it is important that we first cluster evidences $E$ (Line $3$) by content before we apply an iterative removal of each evidence cluster $e^{cl}_i$ (Lines $4$ and $5$).
Each cluster's removal and corresponding counterfactual answer generation is implemented \textit{in parallel} inside the demo for efficiency, as evidences (clusters) can be processed independent of each other.
Moreover, to adjust for non-determinism in LLM generations, we use a Monte Carlo method: in this case it simply means that we repeat the answer generation $m$ times (Lines $6-7$), compute similarities between the answer and the counterfactual answer in each iteration $a^{cf}_{i, j}$ (Line $8$), and average these $m$ values for each cluster.
One minus this average similarity is the \textit{contribution} of each evidence cluster to the original answer: the more similar the answers are, the less an evidence contributed (Line $10$).
These raw answer \textit{contribution scores} $c_i$ for each evidence cluster are then normalized via masked softmax (Line $11$) to derive the final attribution distribution $\mathcal{A}$ (Fig.~\ref{fig:attrib}).
We used the popular DBSCAN algorithm for clustering evidences (Line 3) as it requires only two tunable parameters ($\epsilon$ and $minPts$), and does not require the specification of the number of clusters upfront (hard to predict in our case).
Cosine similarity between the text embeddings
of the answer and the counterfactual answer (in both cases the completed question from the benchmark is prepended for providing the context in which the similarity is sought)
was used in Line 8. Using JinaAI embeddings\footnote{\url{https://huggingface.co/jinaai/jina-embeddings-v3}} to encode evidences which was found to work better than other embedding models (BGE, OpenAI) in this context.

While our algorithm produces fairly accurate results (Table~\ref{tab:attrib}), output distributions (example in Fig.~\ref{fig:attrib}) may sometimes appear to be somewhat degenerate to end users. For example, say among six clusters, the top-evidence cluster is attributed $20\%$ and the remaining five groups are all assigned about $16\%$. We thus introduce a \textit{temperature} parameter $t$ to introduce a higher skew of the distribution towards the top-scoring evidence cluster. 
Answers and counterfactual answers were generated using the cheaper GPT-4o-mini model (instead of GPT-4o) due to budget constraints: every question entails the generation of up to ten counterfactual answers, which entails a substantially higher number of output tokens than generating the intent-explicit questions or the original answers.

\myparagraph{Parameter selection} Parameters $\epsilon$, $minPts$, and $t$ were set to $0.005, 2$, and $0.05$, respectively, using domain and pipeline knowledge. Specifically, (i) the $\epsilon$ parameter in DBSCAN defines the maximum distance between two points for them to be considered neighbors and thus indirectly controls the number of clusters: setting it to be $0.005$ typically resulted in a maximum cluster size of about $2-3$ within $10$ evidences. This is optimal because we cannot attribute answers at a fine granularity if too many evidences get grouped together;
(ii) we work with just ten evidences, so $minPts$, the number of samples in a neighborhood for a point to be considered as a core point, inside the DBSCAN algorithm is set to two;
and (iii) we observed that setting the the temperature to $0.05$ skews the distributions in apparently degenerate cases to $1-3$ evidences: this reflects the average number of gold evidences in \confquestions.

\myparagraph{Differences from \textsc{ContextCite}} Very recently, we came to know about contemporary work on counterfactual attribution in RAG by Cohen-Wang et al.~\cite{cohen2024contextcite} (referred to as \textsc{ContextCite}). Our ideas were independently developed and differ from \textsc{ContextCite} in four major aspects: (i) \textsc{ContextCite} only applies for models with open weights where answer generation likelihoods can be exactly computed, while our idea of estimating evidence attribution scores from similarities of answers and counterfactual answers makes it applicable to RAG systems with closed models as well; (ii) \textsc{ContextCite} does not deal with the problem of redundant information that we tackle via evidence clustering; (iii) \textsc{ContextCite} operates via sentence-level ablations to create counterfactual scenarios: ubiquitous coreferences in sentences then becomes an unsolved concern. We preclude this problem by rather operating at evidence-level, where each evidence is contextualized as discussed earlier, and this makes most evidences relatively self-contained; and (iv) the evaluation of attribution is very different in \textsc{ContextCite}: while they rely on log-probability drops and rank correlations, our benchmark \confquestions allows for more precise evaluation using an accuracy metric.

	
\section{The ConfQuestions Benchmark}
\label{sec:data}

\begin{table} [t] 
 	\newcolumntype{G}{>{\columncolor [gray] {0.90}}c}
 	\resizebox{\columnwidth}{!}{
 	\begin{tabular}{l l}
 		\toprule 
        \textbf{QA}	                                    & \textbf{Statistic}                        \\ \toprule
        \#Conversations	                                & $50$                                      \\
        \#Turns	                                        & $40$ w/ $5$ turns, $10$ w/ $10$ turns     \\
        \#Questions	                                    & $300$ (both in EN and DE)                 \\ \midrule
        Avg. conversational ques. length	            & $9.38$ words                              \\
        Avg. completed ques. length	                    & $14.98$ words                             \\
        Avg. ans. length 	                            & $10.39$ words                             \\ \midrule
        \#Simple questions	                            & $150$                                     \\
        \#Complex questions	                            & $150$                                     \\ \midrule
        \#Questions with answer in passage	            & $100$                                     \\
        \#Questions with answer in list	                & $100$                                     \\
        \#Questions with answer in table	            & $100$                                     \\ \midrule
        \#Conversations with 1 URL	                    & $40$                                      \\
        \#Conversations with 2 URLs	                    & $10$                                      \\
        \#Total URLs in corpus	                        & $215$                                     \\
        \#URLs used for answering	                    & $57$                                      \\ \midrule
        \textbf{Corpus}	                                & \textbf{Statistic}                        \\ \midrule
        \#Spaces	                                    & $10$                                      \\
        \#Pages	                                        & $215$                                     \\ \midrule
        \#Passages	                                    & $2163$                                    \\
        \#Lists	                                        & $1085$                                    \\
        \#Tables	                                    & $110$                                     \\ \midrule
        \#Pages with passages	                        & $215$                                     \\
        \#Pages with lists	                            & $112$                                     \\
        \#Pages with tables	                            & $36$                                      \\ \midrule
        \#Pages with passages and lists	                & $112$                                     \\
        \#Pages with lists and tables	                & $15$                                      \\
        \#Pages with passages and tables	            & $36$                                      \\
        \#Pages with passages and lists and tables      & $15$                                      \\ \midrule
        Median size of passage in words	                & $349$                                     \\
        Median size of list in words	                & $23$                                      \\
        Median size of tables in words	                & $33$                                      \\ \bottomrule
 	\end{tabular}}
 	\caption{Statistics of the \confquestions benchmark.}
 	\label{tab:confquestions}
    \vspace*{-0.3cm}
\end{table}

\myparagraph{Need for a new benchmark} While there are many public QA benchmarks~\cite{roy2022question}, there were none that fit all our desiderata: one that uses heterogeneous pages in enterprise wikis, is suitable for ConvQA with RAG, and contains complex questions in more than one language. So we created our own benchmark \confquestions as follows. We crawled $10$ Spaces under \url{https://openxt.atlassian.net/wiki/spaces}, yielding $215$ public Confluence pages from the developer Atlassian. These contain many documents that are very close to each other content-wise (like test reports for software versions $6-9$, organizational policy over several years, and several meeting notes on a topic), yet each page contains unique information: this is very suitable for evaluating accurate retrieval and generation models. The pages contain a mix of concepts like software compatibility, build configurations, planning, along with mentions of people, dates and software, manifested via heterogeneous elements like tables, lists, and passages. These documents thus nicely simulate an enterprise Wiki space, our focus here.

\myparagraph{Benchmark construction} The authors went through every page to judge the feasibility of generating NL conversations over them. Eventually, $50$ conversations were generated: $80\%$ with 5 turns, $20\%$ longer ones with $10$ turns, making context handling tougher, so $300$ questions in all. $20\%$ conversations also span two URLs instead of one, containing slight topic shifts. The annotators provided questions, intent-explicit versions, answer source (passage/list/table), complexity type (simple/complex), and ground truth answers along with source URLs. The questions were originally generated in English (EN). Native German speakers then translated both conversational and completed questions by hand to German (DE). We took special care to make the benchmark challenging from two aspects: (i) $50\%$ the questions are complex in the sense that they either require joining multiple evidences or involve aggregations, negations, or comparisons; and (ii) the answers to these questions are spread equally across passages, lists, and tables.
So to be able to perform well overall on \confquestions, a RAG systems needs to be able to handle complex questions over heterogeneous elements in multiple languages. Benchmark statistics are in Table~\ref{tab:confquestions}.	
\section{Results and Insights}
\label{sec:results}

\begin{table*} [!t] 
    \vspace*{-0.4cm}
 	\newcolumntype{G}{>{\columncolor [gray] {0.90}}c}
 	\resizebox{\textwidth}{!}{
 	\begin{tabular}{l G G G G G G c c c c c c}
 		\toprule
 	    \textbf{Metric} $\rightarrow$ &	\multicolumn{6}{G}{\textbf{Retrieval precision@1}} &  \multicolumn{6}{c}{\textbf{Answer relevance}}                                 \\ \midrule
        \textbf{Data} $\downarrow$ / \textbf{Contextualization}$^\dagger$ $\rightarrow$                                                                                               &   
        \textbf{NONE}	    & \textbf{+TTL}	& \textbf{+HDR}	& \textbf{+BEF}	& \textbf{+AFT}	& \textbf{+ALL}	& \textbf{NONE}	& \textbf{+TTL}	& \textbf{+HDR}	& \textbf{+BEF}	& \textbf{+AFT}	& \textbf{+ALL}   \\ \toprule
All questions ($600$)	    & $0.398$ & $0.483$ & $0.448$ & $0.453$ & $0.460$ & $\boldsymbol{0.528}$ & $0.388$ & $0.477$ & $0.435$ & $0.445$ & $0.404$ & $\boldsymbol{0.529}$ \\ \midrule
Simple questions ($300$)	& $0.413$ & $0.470$ & $0.477$ & $0.450$ & $0.483$ & $\boldsymbol{0.510}$ & $0.423$ & $0.537$ & $0.482$ & $0.502$ & $0.458$ & $\boldsymbol{0.593}$ \\
Complex questions ($300$)	& $0.383$ & $0.497$ & $0.420$ & $0.457$ & $0.437$ & $\boldsymbol{0.547}$ & $0.352$ & $0.417$ & $0.388$ & $0.388$ & $0.350$ & $\boldsymbol{0.465}$ \\ \midrule
Answer in passage ($200$)	& $0.365$ & $0.410$ & $0.390$ & $0.415$ & $0.420$ & $\boldsymbol{0.445}$ & $0.430$ & $0.547$ & $0.460$ & $0.515$ & $0.500$ & $\boldsymbol{0.603}$ \\
Answer in list ($200$)	    & $0.340$ & $0.475$ & $0.440$ & $0.435$ & $0.490$ & $\boldsymbol{0.560}$ & $0.328$ & $0.422$ & $0.390$ & $0.400$ & $0.340$ & $\boldsymbol{0.507}$ \\
Answer in table ($200$)	    & $0.490$ & $0.565$ & $0.515$ & $0.510$ & $0.470$ & $\boldsymbol{0.580}$ & $0.405$ & $0.460$ & $0.455$ & $0.420$ & $0.372$ & $\boldsymbol{0.477}$ \\ \midrule
English questions ($300$)	& $0.413$ & $0.500$ & $0.467$ & $0.480$ & $0.483$ & $\boldsymbol{0.563}$ & $0.407$ & $0.530$ & $0.472$ & $0.483$ & $0.432$ & $\boldsymbol{0.575}$ \\
German questions ($300$)	& $0.383$ & $0.467$ & $0.430$ & $0.427$ & $0.437$ & $\boldsymbol{0.493}$ & $0.368$ & $0.423$ & $0.398$ & $0.407$ & $0.377$ & $\boldsymbol{0.483}$ \\ \bottomrule
 	\end{tabular}}
    \\ \scriptsize\raggedright $^\dagger$ NONE=No context; TTL = Page title; HDR = Previous heading; BEF = Evidence before; AFT = Evidence after; ALL = All context
 	\caption{Contextualization configurations of \ragonite. The highest value in each row in a group is in \textbf{bold}.}
 	\label{tab:main-res}
    \vspace*{-0.3cm}
\end{table*}

\begin{table} [!t] 
    \vspace*{-0.4cm}
 	\newcolumntype{G}{>{\columncolor [gray] {0.90}}c}
 	\resizebox{\columnwidth}{!}{
 	\begin{tabular}{l G c G}
 		\toprule
        \textbf{Data $\downarrow$} / \textbf{Attribution $\rightarrow$}   & \textbf{Naive}          & \textbf{CFA}              & \textbf{CFA w/ Clusters}  \\ \toprule
        All questions ($364$)	                                          & $0.772$                 & $0.791$                   & $\boldsymbol{0.799}$      \\ \midrule
        Simple questions ($175$)	                                      & $0.771$ 	            & $0.811$                   & $\boldsymbol{0.817}$  \\
        Complex questions ($189$)	                                      & $0.772$                 & $0.772$                   & $\boldsymbol{0.783}$  \\ \midrule
        Answer in passage ($98$)	                                      & $0.816$                 & $\boldsymbol{0.847}$      & $0.806$               \\
        Answer in list ($123$)	                                          & $0.756$                 & $0.780$                   & $\boldsymbol{0.821}$  \\
        Answer in table ($143$)	                                          & $0.755$                 & $0.762$                   & $\boldsymbol{0.776}$  \\ \midrule
        English questions ($192$)	                                      & $\boldsymbol{0.807}$ 	& $0.765$                   & $0.786$               \\
        German questions ($172$)	                                      & $0.733$                 & $\boldsymbol{0.820}$      & $0.814$               \\ \bottomrule
 	\end{tabular}}
    \caption{Accuracies of explanation generation methods.}
 	\label{tab:attrib}
    \vspace{-0.3cm}
\end{table}

\begin{table} [t] 
    \vspace*{-0.4cm}
 	\newcolumntype{G}{>{\columncolor [gray] {0.90}}c}
 	\resizebox{\columnwidth}{!}{
 	\begin{tabular}{l c c c c c c}
 		\toprule
            \textbf{Turns} $\rightarrow$                 & \textbf{1}	        & \textbf{2}	        & \textbf{3}	         & \textbf{4}	         & \textbf{5}              & \textbf{6-10}  \\ \midrule
            Retr. P@1	                                 & $\boldsymbol{0.660}$ & $0.530$	            & $0.500$                & $0.430$	             & $0.490$                 & $0.560$        \\
            Ans. relevance 	                             & $\boldsymbol{0.670}$ & $0.495$	            & $0.465$                & $0.440$	             & $0.555$                 & $0.550$	    \\ \midrule
            \textbf{Corpus} $\rightarrow$                & \textbf{Passages}    & \textbf{Lists}	    & \textbf{Tables}        & \textbf{All}	         &                         &                \\ \toprule
            Retr. P@1	                                 & $0.363$              & $0.422$               & $0.335$                & $\boldsymbol{0.528}$  &                         &                \\ 
            Ans. relevance                               & $0.289$              & $0.407$	            & $0.333$                & $\boldsymbol{0.529}$  &                         &                \\ \midrule
            \textbf{Completion} $\rightarrow$            & \textbf{LLM}	        & \textbf{Human}	    &                        &                       &                         &                \\ \midrule
            Retr. P@1	                                 & $0.528$              & $\boldsymbol{0.658}$  &                        &                       &                         & 	            \\
            Ans. relevance 	                             & $0.529$              & $\boldsymbol{0.627}$  &                        &                       &                         &                \\ \midrule
            \textbf{Linearizer}$^\dagger$ $\rightarrow$  & \textbf{VBL}	        & \textbf{PIPE}	        & \textbf{MD}	         & \textbf{HTML}	     & \textbf{TXT}            & 	            \\ \midrule
            Retr. P@1	                                 & $\boldsymbol{0.528}$ & $0.392$               & $0.382$                & $0.368$	             & $0.372$                 &                \\
            Ans. relevance 	                             & $\boldsymbol{0.529}$ & $0.364$	            & $0.363$                & $0.366$	             & $0.361$                 &	            \\ \midrule
            \textbf{Indexing} $\rightarrow$              & \textbf{Row}	        & \textbf{Table}	    & \textbf{Both}	         &                       &                         &                \\ \midrule
            Retr. P@1	                                 & $0.405$              & $0.492$	            & $\boldsymbol{0.528}$   &                       &                         &                \\
            Ans. relevance	                             & $0.362$              & $0.514$	            & $\boldsymbol{0.529}$   &                       &                         &                \\ \midrule
            \textbf{LLM} $\rightarrow$                   & \textbf{GPT-4o}      & \textbf{Llama3.1}     &                        &                       &                         &                \\ \midrule
            Retr. P@1	                                 & $\boldsymbol{0.528}$ & $0.480$               &                        &                       &                         & 	            \\
            Ans. relevance 	                             & $\boldsymbol{0.529}$ & $0.449$               &                        &                       &                         &                \\ \bottomrule
            \textbf{Embeddings} $\rightarrow$            & \textbf{BGE}         & \textbf{OpenAI}       &                        &                       &                         &                \\ \midrule
            Retr. P@1	                                 & $\boldsymbol{0.528}$ & $0.525$               &                        &                       &                         & 	            \\
            Ans. relevance 	                             & $0.529$              & $\boldsymbol{0.538}$  &                        &                       &                         &                \\ \bottomrule
            \textbf{Ranking} $\rightarrow$               & \textbf{Lexical}     & \textbf{Dense}        & \textbf{Hybrid}        &                       &                         &                \\ \midrule
            Retr. P@1	                                 & $0.430$              & $0.465$               & $\boldsymbol{0.528}$   &                       &                         & 	            \\
            Ans. relevance 	                             & $0.417$              & $0.443$               & $\boldsymbol{0.529}$   &                       &                         &                \\ \bottomrule        
            \textbf{Reranking} $\rightarrow$             & \textbf{BGE+RRF}     & \textbf{RRF}          &                        &                       &                         &                \\ \midrule
            Retr. P@1	                                 & $\boldsymbol{0.528}$ & $0.417$               &                        &                       &                         & 	            \\
            Ans. relevance 	                             & $\boldsymbol{0.529}$ & $0.507$               &                        &                       &                         &                \\ \bottomrule 
 	\end{tabular}}
    \\ \scriptsize \raggedright $^\dagger$ VBL=Verbalization; PIPE = Piped; MD = Markdown; HTML = HTML format; TXT = Plaintext
 	\caption{\ragonite ablations. Highest values in rows in \textbf{bold}.}
 	\label{tab:analysis}
    \vspace*{-0.3cm}
\end{table}

\subsection{Metrics}
\label{subsec:metrics}

We use \textit{Precision@1} as our retrieval metric (since there is usually one gold URL), and \textit{Answer relevance} as judged by GPT-4o ($0$ for non-relevant, $0.5$ for partially relevant, $1$ for relevant) against gold answers, as our metric for answer quality. Question-wise results are averaged over the full dataset. We use \textit{accuracy} for evaluating attribution explanations as follows: we find the evidence that has the highest attribution score (for a cluster, the score of the cluster is considered to be the score of each evidence inside), and find its source URL. If this URL matches that of the gold answer, then accuracy is $1$, else $0$. The average accuracy over all questions is then reported. This is interpretable as a percentage if multiplied by $100$.

For both Precision@1 and Attribution accuracy, we match the URL of the top-evidence against the answer URL in the benchmark. This is a proxy for ground truth: a perfect case would be to have gold labels at evidence-level, but this is very difficult to annotate for our benchmark, as gold answers are often composed of nuggets of information spread over the entire document.

\subsection{Setup}
\label{subsec:setup}

We now run and evaluate \ragonite in various configurations over the $600$ English and German questions in \confquestions. The proposed default configuration uses: (i) all available context; (ii) all evidences as the corpus; (iii) LLM-generated intent-explicit questions; (iv) verbalization as the linearization technique for table records; (v) both row- and table-level indexing for table elements; (vi) GPT-4o as our LLM of choice; (vii) BGE as our text embedding model; (viii) hybrid search as our retrieval mode; and (ix) BGE+RRF reranking over the hybrid search results.
We feed the reranked top-10 evidences ($k=10$) by the retriever into the LLM.
Contextualization and attribution variants are discussed in Sec.~\ref{subsec:key}: all other ablation analysis is postponed to Sec.~\ref{subsec:analysis}. 

\begin{figure} [t]
	\centering
	\includegraphics[width=\columnwidth]{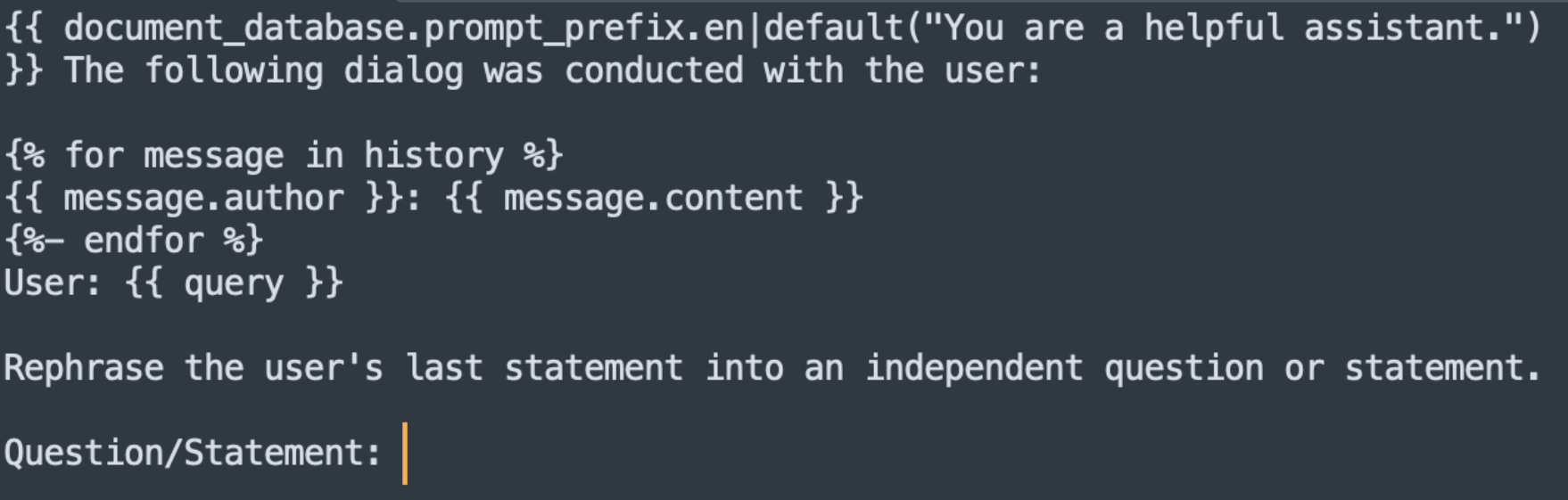}
    \vspace*{-0.7cm}
	\caption{LLM prompt for rephrasing a conversational question into an intent-explicit form.}
	\label{fig:prompt-1}
    \vspace*{-0.3cm}
\end{figure}

\begin{figure} [t]
	\centering
	\includegraphics[width=\columnwidth]{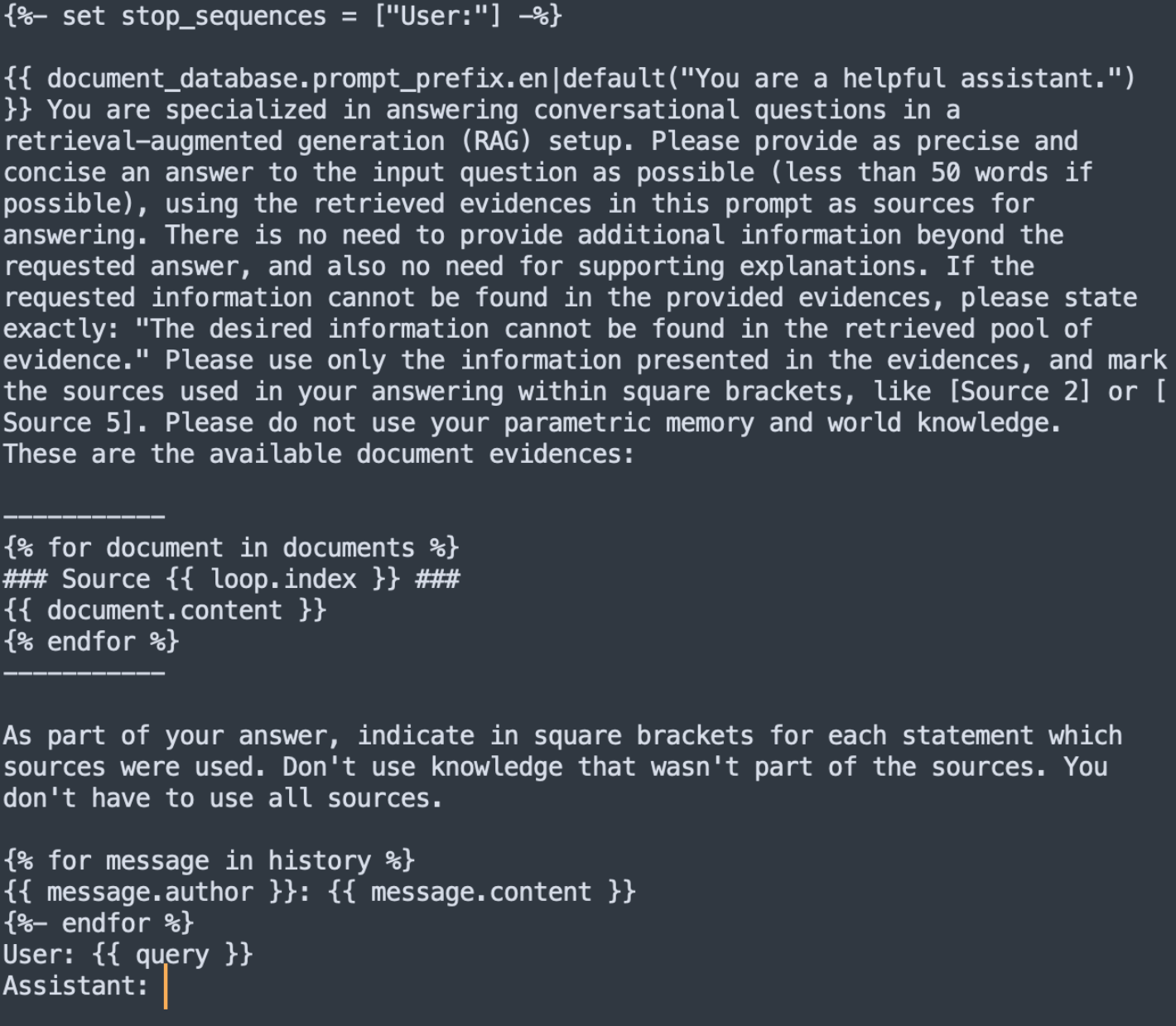}
    \vspace*{-0.7cm}
	\caption{LLM prompt for generating an answer from the retrieved pool of evidence.}
	\label{fig:prompt-2}
\end{figure}

\begin{figure} [t]
	\centering
	\includegraphics[width=\columnwidth]{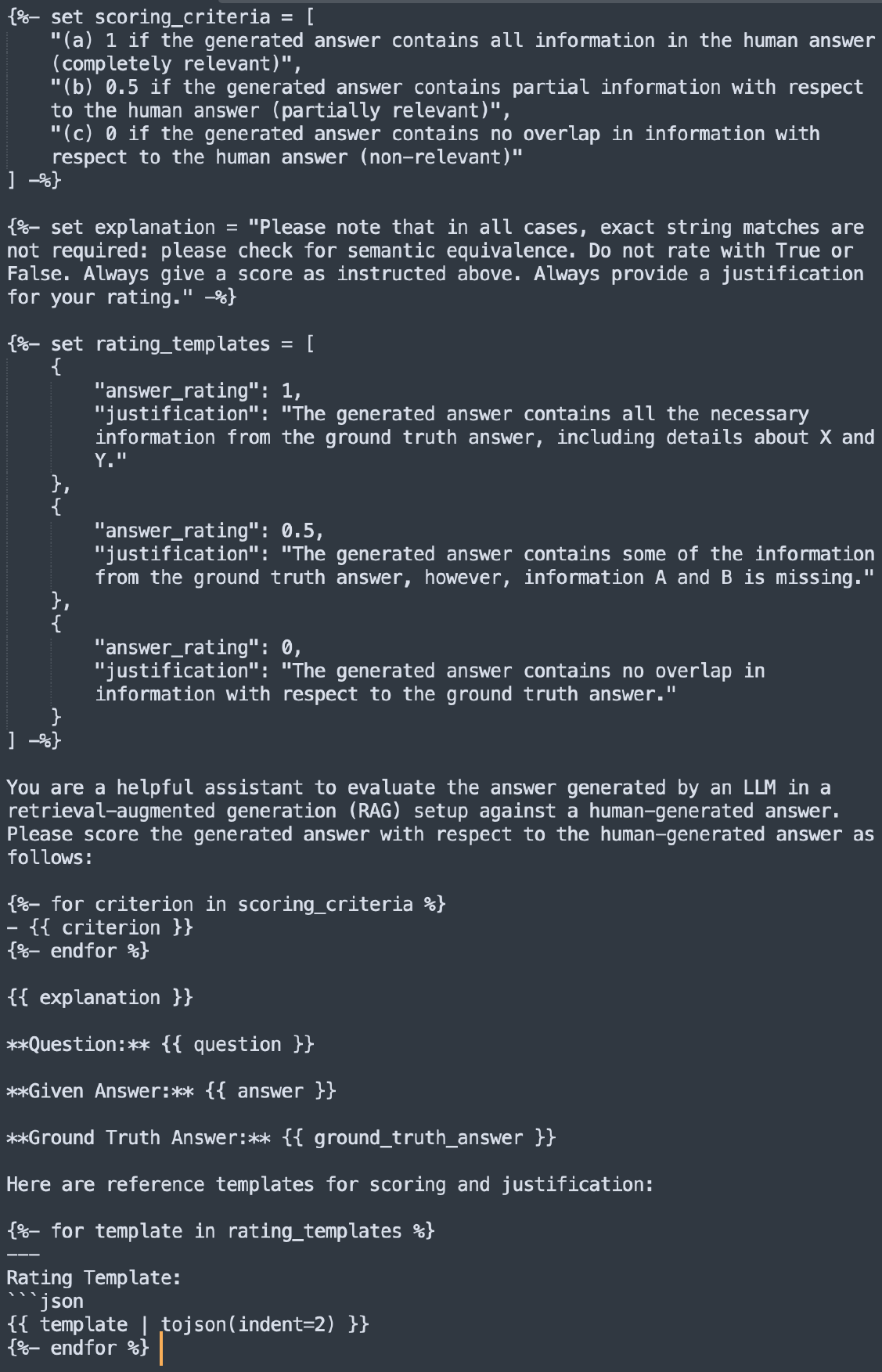}
    \vspace*{-0.7cm}
	\caption{LLM prompt for evaluating a generated answer according to the Answer Quality measure.}
	\label{fig:prompt-3}
\end{figure}

\myparagraph{Prompts} There are three prompts used in \ragonite: (i) for rephrasing a conversational question; (ii) for generating an answer from retrieved evidences; and (iii) for evaluating a generated answer. These are presented in Figs.~\ref{fig:prompt-1},~\ref{fig:prompt-2}, and~\ref{fig:prompt-3}, respectively (used for both English and German questions).

\subsection{Key findings}
\label{subsec:key}

\myparagraph{Evidence contextualization improves performance} Contextualization results are in Table~\ref{tab:main-res}. We make the following observations:
(i) Adding all the suggested contexts (ALL) consistently led to significantly improved performances for retrieval and generation. 
This means that enhancing evidences from documents with surrounding context is generally beneficial for RAG over collaborative wikis: content authors are often efficiency-oriented and do not repeat information inside passages, lists, or tables that are already mentioned in page titles or previous headings, for example. Similarly, adding preceding and succeeding texts, give coherence to the evidence and help in better retrieval (users often use localizing words in questions that may not lie exactly inside the pertinent evidence but could be nearby in the document). Surrounding texts clearly also help the LLM generate more consistent and relevant responses; (ii) Systematically, the page title (TTL) was found to be the single most helpful context to be added to the evidences ($0.483$ P@1 for TTL vs. $0.528$ for ALL, and $0.477$ Answer Relevance for TTL vs. $0.529$ for ALL), with each of the other three individual add-ons bringing in substantial improvements over NONE (ALL > TTL > AFT > BEF > HDR > NONE for P@1 and ALL > TTL > BEF > HDR > AFT > NONE for answer relevance).
No feature hurts performance.
It is notable that succeeding content also helps (a notable case is that of table footers), while common coreferencing methods only contextualize using preceding text;
(iii) These improvements are systematic with respect to question complexity (simple/complex), answer source (passage/list/table), and language (English/German), demonstrating that contextualization is a worthwhile operation across diverse question types. We also note that the biggest gains in retrieval ($0.220$ gain in P@1 of ALL over NONE) and answering ($0.179$ gain in answer relevance of ALL over NONE) performance are achieved for answer-in-list questions -- a very common QA scenario for enterprise wikis where one asks for meeting notes, attendees and action items -- all often stored as lists.

\myparagraph{Counterfactual attribution is effective} While counterfactual evidence has been proven to be a tool for discovering causal factors~\cite{pearl2020book}, our particular implementation still needed proof of efficacy. Results in Table~\ref{tab:attrib} (Column $4$) show that we consistently reach accuracies close to $80\%$, definitely an acceptable number in terms of user trust in \ragonite explanations.
The numbers in parentheses in Column $1$ show over how many questions the respective averages were computed in Columns $2-4$. For questions where the gold document/evidence was not retrieved in the top-$10$, there is no chance that the attribution module could spot the correct evidence. Hence these are removed from this evaluation and we do not see round numbers like $600/200/100$ as in Table~\ref{tab:main-res}.
Looking at individual data slices across rows, it is satisfying to observe that the high performance was not due to skewed success on easier slices: improvements over standard or naive attribution are quite systematic over almost all criteria (varying only within a narrow span of $77-82\%$).
In the naive mode (also a baseline in~\cite{cohen2024contextcite}), we compute cosine similarities between the answer and the evidences, and then use softmax scores over this distribution as the attribution scores for the corresponding evidences. The top scoring evidence is used to compute accuracy (Sec.~\ref{subsec:metrics}).
Within the counterfactual approaches, the one with clustering generally does better (with-clusters best in $5/8$ cases, without-clusters in $2/8$: marked in boldface in Table~\ref{tab:attrib}).
Understandably, complex questions and answer-in-table questions are slightly more difficult cases to solve (relatively lower accuracies of $\simeq 77-78\%$). Performance for German was found to be slightly better than that for English.

\subsection{In-depth RAG analysis}
\label{subsec:analysis}

In Table~\ref{tab:analysis}, we report several drill-down experiments with \ragonite configurations. Whenever one configuration choice was altered, the remaining values were held constant as per our default configuration specified in Sec.~\ref{subsec:setup}.
The study leading to this table was also used to \textit{select} our default configuration. Specific observations are listed below:
\squishlist
    \item \textbf{[Conversation turns]} In Rows $1-3$, we observe that while there is an understandable drop after the first turn owing to intent-implicit questions, \ragonite remains fairly consistent in performance in deeper turns (default configuration used).
    \item \textbf{[Answer source]} In Rows $4-6$, we see clear proof that heterogeneous evidence helps QA: the ability of tapping into a mixture of sources notably improves RAG performance.
    \item \textbf{[Question completion]} In Rows $7-9$, we note that accurate question completion is still a bottleneck, as using human completions from the benchmark substantially improves metrics.
    \item \textbf{[Table linearization]} In Rows $10-12$, verbalization of table records indeed makes table contents 
    more retrievable by a hybrid retriever (see below) as well as more digestible by an LLM, compared to natural linearization alternatives (the piped mode can be found in~\cite{wang2024chain,li2023table}; markdown simply refers to the table in markdown format\footnote{\url{https://www.markdownguide.org/extended-syntax/\#tables}};
    HTML refers to the raw HTML of the table; plaintext is the remaining text of the table after stripping all HTML tags and attributes). But we also prefer verbalizations in the \ragonite interface owing to its scrutability by an end-user.
    \item \textbf{[Table indexing]} In Rows $13-15$, we find that retaining both table embeddings and individual row embeddings in the DB is preferable to having only either of these. This is understandable in the sense that some questions need an aggregation over several rows or the entire table, this is difficult if the full table is not indexed (and retrieved). On the other hand, questions pertaining to specific cells or joins concerning $1-2$ rows may be harder to fish out from very large tables by an LLM: here the ability to retrieve individual rows is helpful.
    \item \textbf{[Generation LLM]} In Rows $16-18$, we find that using GPT-4o led to superior answer quality than Llama3.1-8B.
    Unfortunately the presence of large lists and table evidences in Confluence often led to the context window being exceeded for Llama. The same LLM was used for both completed question and final answer generation. The P@1 is also lower for Llama ($0.480$ vs. $0.528$ for GPT-4o) because the retrieval is carried out via the intent-explicit question generated by Llama (slightly inferior to corresponding completions from GPT-4o). As requested in the prompt, the out-of-scope (OOS) message ``The desired information cannot be found in the retrieved pool of evidence.'' was triggered $199/600$ times ($33.2\%$) for the default \ragonite configuration with GPT-4o. Computing the Precision@10 to be $60.7\%$, an OOS  message would be expected $39.3\%$ of the times. Since this is very close to the actual measured rate of the OOS message, we can say that GPT-4o showed high instruction-following capabilities and did not use its parametric knowledge. 
    \item \textbf{[Embedding model]} In Rows $19-21$, we see that using \texttt{bge-m3} embeddings
    for encoding text led to marginally better retrieval (P@1$=0.528$) than the recent \texttt{text-embedding-3-small} model
    from OpenAI (P@1$=0.525$). While using the OpenAI model led to slightly better answer quality, better retrieval is the primary purpose of embeddings, leading to our choice of BGE as default. Moreover, BGE is used via local deployment, that we find preferable to the OpenaI API due to latency and cost issues.
    \item \textbf{[Evidence ranking]} In Rows $22-24$, hybrid search was found to be superior than lexical or dense search built inside ChromaDB. We used 10 evidences each from lexical and dense search. 
    \item \textbf{[Evidence reranking]} In Rows $25-27$, reranking evidences in the common pool of lexical and dense retrieval via the multilingual BGE reranking model+RRF
    to create the hybrid ranked list, was found to be better than using only RRF. In RRF, the final list is created by ordering via a new evidence score that is the sum of the reciprocals of the ranks of the evidence in the lexical and dense rankings (a parameter $k=60$ is added to the original rank~\cite{cormack2009reciprocal} before computing the reciprocal).
\squishend

\section{Conclusion and future work}
\label{sec:confut}

\ragonite is an explainable RAG pipeline, that makes every intermediate step open to scrutiny by an end user. We propose flexible strategies for adding context to heterogeneous evidence, and causally explain LLM answers using counterfactual attribution.
Future work 
will introduce cost and time-efficient agentic workflows.

\section*{Acknowledgements}

This research was funded by the German Federal Ministry for Economic Affairs and Climate Action (BMWK) through the project OpenGPT-X (project no. 68GX21007D).
We thank Paramita Mirza from the NLP team at Fraunhofer IIS for useful comments on an initial version of this manuscript, 
and acknowledge Sabrina Stehwien, Md. Saiful Islam, Darina Gold and Viktor Haag from the same team for useful inputs at various stages of this work.

\bibliographystyle{ACM-Reference-Format}
\bibliography{2025-arxiv-dp-ragonite}

\end{document}